\documentclass[letterpaper]{article} 
\usepackage{aaai2027}  
\usepackage[hyphens]{url}  
\usepackage{graphicx} 
\usepackage{natbib}  
\usepackage{caption} 
\usepackage{booktabs}
\usepackage{tikz}
\usetikzlibrary{arrows.meta,positioning,fit,backgrounds}

\title{Annotrace: A Workbench for Step-Level Annotation of Agent Traces}

\author{
    Chenyi Huang\textsuperscript{\rm 1},
    Arun Arokiaraj\textsuperscript{\rm 1}\corresponding,
    Bin Wu\textsuperscript{\rm 1,2},
    Sahan Bulathwela \textsuperscript{\rm 1,2} \corresponding
}
\affiliations{
    \textsuperscript{\rm 1}Department of Computer Science, University College London, London, United Kingdom\\
    \textsuperscript{\rm 2}Centre for Artificial Intelligence, University College London, United Kingdom\\
    \texttt{\{chenyi.huang.25, m.bulathwela\}@ucl.ac.uk}
}

\begin{document}

\maketitle

\begin{abstract}
Every study of why LLM agents fail begins by labelling traces by hand, and
every such study builds its own throwaway tool. Observability platforms record
agent traces but offer no structured labelling; annotation tools label text but
have no notion of a step, a delegation or a tool call, creating uncertainty around the smallest atomic unit of execution one should focus on. Annotrace reads six
trace formats, including OpenTelemetry GenAI, LangSmith and Langfuse, and turns
each run into an ordered, nested list of steps. The annotator writes their own
labelling codebook;
a frontier model proposes a starting set from a sample of their own traces,
answers first, and is asked exactly the questions the human is asked shifting towards a synergistic human-AI teamed annotation process
. Annotrace
further measures agreement between any two annotator agents, human or AI. On a multi-agent
question-answering corpus it reproduces a previous manual effort's 235
annotation units exactly, and two models labelling the same traces agree 91\%
of the time at $\kappa = 0.82$.
\end{abstract}

\section{Introduction and Problem}

\begingroup
\centering
\begin{tikzpicture}[
  font=\scriptsize,
  who/.style={anchor=west, font=\scriptsize\itshape, text=black!60, inner sep=0pt},
  did/.style={anchor=west, align=left, inner sep=0pt},
  chip/.style={draw, rounded corners=1.2pt, minimum width=0.5cm,
               minimum height=0.25cm, inner sep=0.8pt, anchor=east},
  no/.style={chip, fill=red!8, draw=red!35},
  yes/.style={chip, fill=green!8, draw=green!40},
]
\newcommand{\stprow}[6]{%
  \node[who] at (#1,#2) {#3};
  \node[did] at (#1+1.55,#2) {#4};
  \node[#5] at (8.0,#2) {#6};}

\node[anchor=west, font=\scriptsize\bfseries] at (0,0.78)
  {Outcome: correct \normalfont\itshape--- the run's only label};
\node[anchor=west, font=\scriptsize, text=black!60] at (0,0.44)
  {a coordinator delegates sub-questions to workers \textit{e1} and \textit{e2}};
\draw[black!25] (0,0.24) -- (8.0,0.24);

\stprow{0}{0}{coordinator}{asks ``notable American character actors?''}{no}{No}
\stprow{0.34}{-0.36}{worker e2}{Charlie Chaplin, after four search cycles}{no}{No}
\stprow{0}{-0.72}{coordinator}{asks the same question again}{no}{No}
\stprow{0.34}{-1.08}{worker e2}{Marlon Brando, after four more cycles}{no}{No}
\stprow{0}{-1.44}{coordinator}{asks ``who owns KWPW?''}{yes}{Yes}
\stprow{0.34}{-1.80}{worker e1}{Bill McCutcheon, two cycles}{yes}{Yes}
\stprow{0}{-2.16}{coordinator}{asks whether he is an actor}{yes}{Yes}
\stprow{0.68}{-2.52}{cycle 3}{reasons ``software engineer'', answers ``yes''}{no}{No}
\stprow{0}{-2.88}{final answer}{Bill McCutcheon}{yes}{Yes}

\draw[black!25] (0,-3.08) -- (8.0,-3.08);
\end{tikzpicture}
\captionof{figure}{One multi-agent run on HotpotQA \citep{yang2018hotpotqa}. It answers
correctly, so a trace-level scheme calls it a success. Its steps record eight
wasted search cycles and a verification the worker's own reasoning
contradicts. Indentation is delegation depth; the decisive fault sits inside a
recovered cycle (Figure~\ref{fig:tar}).}
\label{fig:why}
\endgroup

Understanding the behavior of LLM agents requires looking beyond whether a task ultimately succeeds or fails to what happens during execution. Agent trajectories expose the intermediate decisions, tool interactions, and delegations through which an outcome is produced. A successful run may still contain incorrect or redundant steps, while a failure may originate much earlier than the final output (Figure~\ref{fig:why}). Consequently, recent work on agent diagnosis increasingly relies on manually inspecting and annotating execution traces: MAST derives failure modes from annotated trajectories \citep{cemri2025mast}; Who\&When annotates failure logs for agent and step-level attribution \citep{zhang2025whichagent}; AgentErrorTaxonomy similarly studies where agents fail \citep{zhu2025where}; and AgentRx checks constraints against execution trajectories \citep{barke2026agentrx}. Yet such annotation is typically tied to a particular analysis objective and trace format, limiting reuse across studies.

Existing tools address only parts of this problem. Observability platforms such as LangSmith, Langfuse, and Arize Phoenix record and display agent traces, with OpenTelemetry providing emerging conventions for their representation \citep{otel2025genai}, but they are not designed for structured human annotation. Conversely, general-purpose annotation tools such as Label Studio \citep{labelstudio}, doccano \citep{doccano}, and INCEpTION \citep{inception} provide annotation workflows over documents or text spans. Agent trajectories are structurally different: they contain heterogeneous and potentially nested execution units, including decisions, delegations, tool calls, and responses. Moreover, different studies ask different questions of these steps and often construct their own annotation schemes and procedures. Trajectory annotation therefore requires not only preserving execution structure, but also flexibly defining what to annotate and supporting the annotator throughout this process.

We present \textbf{Annotrace}, a workbench for structured, step-level annotation of agent trajectories. Annotrace (1) converts heterogeneous trace formats into a unified nested representation, making individual execution steps directly annotatable; (2) allows users to define what to annotate through editable questions and labels rather than a fixed taxonomy; and (3) integrates LLMs into the annotation workflow to propose schemas and pre-annotate traces using the same questions presented to human annotators. Together, Annotrace provides an end-to-end workflow from raw agent traces to structured step-level annotations, while keeping the annotation objective and final labels under the user's control.

\begin{figure}[t]
\centering
\resizebox{0.8\columnwidth}{!}{\begin{tikzpicture}[
  font=\scriptsize,
  fmt/.style={draw, rounded corners=1pt, minimum height=0.34cm,
              text width=2.0cm, align=center, inner sep=1pt},
  bar/.style={draw, fill=gray!10, minimum height=2.3cm, text width=0.40cm,
              align=center, inner sep=1pt},
  stp/.style={draw, rounded corners=1pt, minimum height=0.28cm,
              text width=1.15cm, align=center, inner sep=1pt, font=\tiny},
  act/.style={draw, rounded corners=1pt, minimum height=0.40cm,
              text width=1.50cm, align=center, inner sep=1.5pt},
  mdl/.style={act, fill=gray!14},
  arr/.style={-{Stealth[length=1.2mm]}, thin, gray!65},
  tree/.style={gray!55, thin},
]
\node[fmt] (f1) at (0.92, 1.10) {OpenTelemetry};
\node[fmt] (f2) at (0.92, 0.66) {LangSmith};
\node[fmt] (f3) at (0.92, 0.22) {Langfuse};
\node[fmt] (f4) at (0.92,-0.22) {chat messages};
\node[fmt] (f5) at (0.92,-0.66) {Who\&When};
\node[fmt] (f6) at (0.92,-1.10) {coordinator};

\node[bar] (ad) at (2.50, 0) {\rotatebox{90}{adapters}};
\foreach \i in {1,...,6} { \draw[arr] (f\i.east) -- (ad.west); }

\node[draw, dashed, gray!55, rounded corners=2pt,
      minimum width=3.7cm, minimum height=2.3cm] (box) at (5.35,0) {};
\node[font=\scriptsize] at (5.35,1.42) {one nested step model};
\node[stp] (s1) at (4.20, 0.66) {decision};
\node[stp] (s2) at (4.85, 0.22) {tool call};
\node[stp] (s3) at (4.85,-0.22) {reply};
\node[stp] (s4) at (5.50,-0.66) {cycle};
\draw[arr] (ad.east) -- (box.west);
\draw[tree] (3.78, 0.52) -- (3.78,-0.22) ;
\draw[tree] (3.78, 0.22) -- (4.27, 0.22);
\draw[tree] (3.78,-0.22) -- (4.27,-0.22);
\draw[tree] (4.43,-0.36) -- (4.43,-0.66) -- (4.92,-0.66);

\node[mdl] (mp) at (0.92,-2.30) {model proposes};
\node[act] (q)  at (2.85,-2.30) {your questions};
\node[act] (an) at (4.78,-2.30) {annotate};
\node[mdl] (ma) at (6.71,-2.30) {model answers};
\draw[arr] (mp) -- (q);
\draw[arr] (q)  -- (an);
\draw[arr] (an) -- (ma);
\draw[arr] (box.south) -- ++(0,-0.42) -| (mp.north);
\node[font=\scriptsize] at (4.10,-2.95)
  {one object writes the questions \emph{and} the model's instructions};
\end{tikzpicture}}

\vspace{4pt}

\begin{tikzpicture}[
  font=\scriptsize,
  raw/.style={draw, align=left, text width=2.75cm, fill=gray!6, inner sep=2.5pt,
              font=\ttfamily\scriptsize},
  cyc/.style={draw, rounded corners=1.5pt, align=center, minimum height=0.30cm,
              text width=0.62cm, fill=gray!18, inner sep=1pt},
  arr2/.style={-{Stealth[length=1.3mm]}, thin}
]
\node[raw] (r) at (0,0)
  {<think>\\<search>\\<information>\\<think>\\<search>\\<answer>};
\node[font=\scriptsize, above=1pt of r] {one text field};
\node[cyc] (t1) at (2.65,0.38) {T};
\node[cyc] (a1) at (3.42,0.38) {A};
\node[cyc] (r1) at (4.19,0.38) {R};
\node[cyc] (t2) at (2.65,-0.32) {T};
\node[cyc] (a2) at (3.42,-0.32) {A};
\node[cyc] (r2) at (4.19,-0.32) {R};
\node[font=\scriptsize, above=1pt of a1] {two labelled cycles};
\draw[arr2] (r.east) -- ++(0.28,0) |- (t1.west);
\draw[arr2] (r.east) -- ++(0.28,0) |- (t2.west);
\draw[arr2, gray!60] (t1) -- (a1); \draw[arr2, gray!60] (a1) -- (r1);
\draw[arr2, gray!60] (t2) -- (a2); \draw[arr2, gray!60] (a2) -- (r2);
\end{tikzpicture}
\caption{Annotrace workflow. Top: adapters convert six trace formats into
nested steps. Middle: a model proposes an annotation schema and answers the
same questions as the human annotator. Bottom: adapters can recover and
separately label Thinking--Action--Response cycles embedded in a single
text field.}
\label{fig:arch}
\label{fig:tar}
\end{figure}

\section{Technical Details and Architecture}


\paragraph{Adapters.}
An adapter turns one format into \texttt{Step} objects carrying named fields,
an actor, a kind and a nesting depth (Table~\ref{tab:formats}).
Detection reads file content --- each adapter looks for a marker no other
format uses, such as \texttt{run\_type} with \texttt{dotted\_order} for
LangSmith --- so the user need not know what they were given.
Granularity differs by format and is exposed rather than chosen. Where a
worker's reasoning is one text field of interleaved \texttt{think},
\texttt{search}, \texttt{information} and \texttt{answer} tags, the adapter
recovers Thinking-Action-Response cycles (Figure~\ref{fig:tar}). A step
containing finer steps is left unselected, so nothing is labelled twice.

\paragraph{Schema as questions.}
A schema is a list of questions, each with a prompt, the labels the user typed,
the step kinds it applies to, and optionally a condition on an earlier answer:
\emph{ask ``what went wrong?'' only when ``severity'' is ``Critical''}.
Conditions are authored in the interface, not a file. No taxonomy ships; the
three included schemas are examples a user may delete.

\paragraph{The model in the loop.}
A model is used twice, both times asked the questions the human is asked,
generated from the same object. It reads ten randomly sampled runs and proposes
a starting schema (at most three questions, four labels each, with house
rules), then answers a trace. Every answer records whether a person or the
model produced it; a suggestion never overwrites a human decision, and editing
one makes it the annotator's. Any of 46 models may be chosen, so a cheap model
can draft and an expensive one check.

\paragraph{Agreement.}
Two annotation sets are compared question by question, reporting raw agreement,
Cohen's $\kappa$ and the steps they disagree on; either side may be the open
session. $\kappa$ is reported because raw agreement flatters an uneven schema.

\begin{table}[t]
\centering
\setlength{\tabcolsep}{4pt}
\resizebox{0.8\linewidth}{!}{\begin{tabular}{@{}llc@{}}
\toprule
\textbf{Format} & \textbf{Step kinds} & \textbf{Nested} \\
\midrule
OpenTelemetry & invoke\_agent, chat, execute\_tool & yes \\
LangSmith     & chain, llm, tool, retriever        & yes \\
Langfuse      & agent, generation, tool, span      & yes \\
Chat messages & message, tool call, tool result    & no  \\
Who\&When     & message                            & no  \\
Coordinator   & decision, reply, cycle, output     & yes \\
\bottomrule
\end{tabular}}
\caption{The six adapters. Kinds are each format's own vocabulary; a question
can be restricted to any subset of them.}
\label{tab:formats}
\end{table}

\section{Innovation and Significance}

\paragraph{The schema comes from the user's data.}
Existing tools ask for a label set before the annotator has read anything.
Annotrace shows the traces first, then proposes labels from a sample of them.
Given the one-line goal ``which step caused the run to give a wrong answer'', a
model proposed \emph{Was this step correct? Yes / No / Irrelevant} with rules
including ``pick No only if you can see a clear mistake''. The proposal lands
in editable fields and is never applied unasked.

\begin{figure}[!t]
\centering
\includegraphics[width=0.7\columnwidth,height=6.5cm,keepaspectratio]{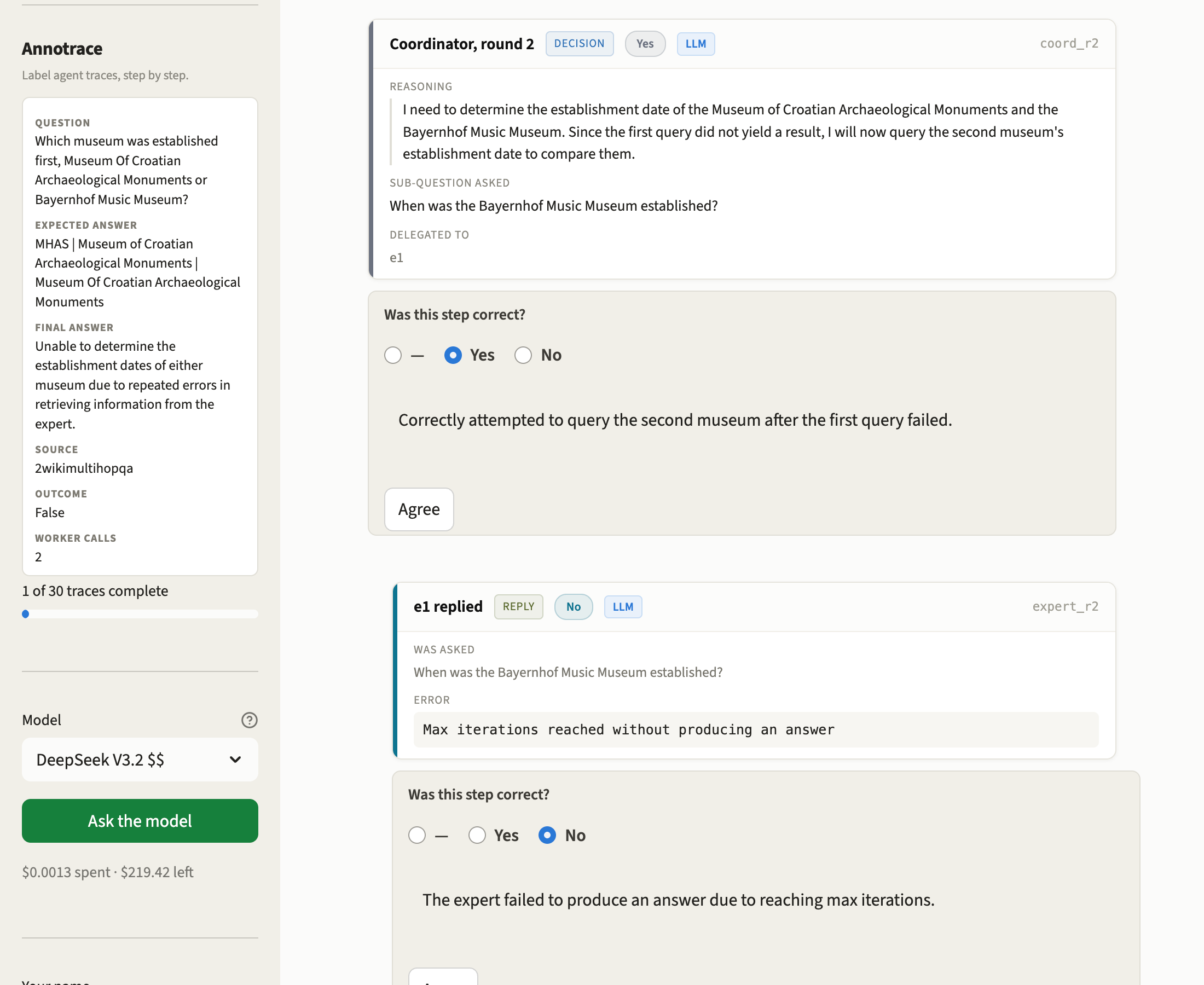}
\caption{The annotation screen. The sidebar holds the run's question and the
agent's answer, so both stay in view while the steps scroll. Colour marks the
kind of step until a label is chosen, then marks the label. Steps still wanting
an answer say so.}
\label{fig:gui}
\end{figure}

\paragraph{Evidence.}
On a 40-run multi-agent corpus the coordinator adapter reproduced all 235
Thinking-Action-Response unit keys of a previous hand-built effort, on 40 of 40
runs, so earlier work still aligns. Two models labelling the same six runs
agreed on 91\% of 22 answers at $\kappa = 0.82$. A schema proposal takes about
four seconds, annotating a run three to five, at a fraction of a cent per call.
Renaming a label keeps and marks the answers that used the old name rather than
discarding them. Figure~\ref{fig:gui} shows the annotation screen: a visitor
loads a run, types one line saying what they are looking for, takes the
proposed labels, and disagrees with the model on a step, in under two minutes.

\end{document}